\documentclass[journal]{IEEEtran}
\usepackage{graphicx}
\usepackage{subfigure}
\usepackage{cite}
\usepackage{amssymb, amsmath}
\usepackage{multirow}
\usepackage{color}

\begin{document}

\title{Multimodal LLM Integrated Semantic Communications for 6G Immersive Experiences}

\author{Yusong~Zhang, Yuxuan~Sun,~\IEEEmembership{Member,~IEEE,} Lei~Guo, Wei Chen,~\IEEEmembership{Senior Member,~IEEE,} 
\\Bo Ai,~\IEEEmembership{Fellow,~IEEE,}
Deniz G{\"u}nd{\"u}z,~\IEEEmembership{Fellow,~IEEE}
\thanks{
Y. Zhang, Y. Sun (Corresponding Author), L. Guo, W. Chen, B. Ai are with School of Electronic and Information Engineering, Beijing Jiaotong University, Beijing 100044, China.

D. G{\"u}nd{\"u}z is with the School of Electrical and Electronic Engineering, Imperial College London, London SW7 2BT, U.K.
%Corresponding author.
%This work is sponsored in part by ...
}
%\thanks{This work is supported by the Natural Science Foundation of China (62301024, U2468201).%, and the Young Elite Scientists Sponsorship Program by CAST.
%}
}

\maketitle

\begin{abstract}
6G networks promise revolutionary immersive communication experiences including augmented reality (AR), virtual reality (VR), and holographic communications. These applications demand high-dimensional multimodal data transmission and intelligent data processing in real-time, which is extremely challenging over resource-limited wireless communication systems. Moreover, a joint understanding of the environment, context, and user intent is essential to deliver task-relevant content effectively.
This article presents a novel multimodal large language model (MLLM) integrated semantic communications framework, termed MLLM-SC, which fully leverages reasoning and generative capabilities of pre-trained foundation models for context-aware and task-oriented wireless communication. The MLLM-SC framework adopts a device-edge collaborative architecture. At the edge, MLLM-empowered semantic guidance module analyzes multimodal inputs, user intents, and channel conditions to generate importance-aware attention maps prioritizing semantically critical information. An importance-aware semantic encoder and a resource-adaptive semantic decoder are jointly designed and optimized, which can utilize the semantic guidance for adaptive bandwidth allocation and high-quality content reconstruction or generation. Extensive case studies on visual question answering for AR/VR applications and diffusion-driven image generation validate the effectiveness of MLLM-SC.
%The MLLM-SC framework comprises three core components within a device-edge collaborative architecture. First, an MLLM-empowered semantic guidance module analyzes multimodal inputs, user intents, and channel conditions to generate importance-aware attention maps prioritizing semantically critical information. Second, an importance-aware semantic encoder employs dual-path architecture with cross-attention mechanisms to dynamically allocate transmission resources based on semantic guidance and channel states. Third, a resource-adaptive semantic decoder integrates variational autoencoders and conditional diffusion models for high-quality content reconstruction and generation under varying constraints.The framework demonstrates real-time environment understanding, importance-aware transmission, highly-efficient communication through semantic compression, and channel-adaptive performance.
%showing significant improvements in pixel-level reconstruction quality and semantic preservation compared to conventional approaches. The MLLM-SC framework represents a paradigm shift toward intelligent semantic communications, enabling efficient immersive experiences in resource-constrained 6G environments.
\end{abstract}

%IEEE Communications Magazine Submission Policy: \textbf{5500 words} in total (inclusive of title, authors’ names/info, abstract, body (introduction to conclusion), figures, tables, captions, footnotes, acknowledgements, references, authors’ bios(<150 words each), etc.).\textbf{ 6 figs/tables.}  \textbf{No equations} (In justified cases, up to three simple equations may be allowed. ) \textbf{15 archival references }(complete with publication name, year, volume, issue and page numbers is recommended. Authors’ Self-citation must be minimized. )

\section{Introduction}

6G is expected to support immersive communication experiences that seamlessly integrate the physical and digital worlds \cite{shen2023toward}. Envisioned applications such as augmented reality (AR), virtual reality (VR), mixed reality (XR), and holographic communications demand real-time transmission and processing of massive multimodal data including high-resolution videos or images, dense point clouds, audio streams, and sensor information. However, bandwidth-constrained wireless channels, along with limited computing power and storage capacity of devices, pose significant challenges to high-dimensional data transmission and intelligent data processing.

Powered by artificial intelligence (AI), semantic communications is an emerging paradigm that focuses on transmitting different features of data, rather than merely reconstructing the raw input signal \cite{Gunduz2023Beyond}. An edge–device collaborative architecture is commonly adopted, with advanced neural networks deployed at edge servers to enhance semantic extraction and processing. Deep learning-based joint source–channel coding (DJSCC) is a key approach in semantic communications, which optimizes information compression and transmission processes jointly in an end-to-end fashion \cite{Jialong-ComMag}. Recent advances in generative AI, particularly variational autoencoders, generative adversarial network (GANs) and diffusion models, have further enhanced semantic communications by enabling high-quality reconstruction and generation from compressed semantic representations \cite{Erdemir2023Generative}. However, existing semantic communication systems often lack comprehensive understanding of complex environments and diverse user intents, and thus may have difficulty extracting the most task-relevant semantic information.

The emergence of large language models (LLMs), especially multimodal LLMs (MLLMs), opens new avenues to overcome the above challenges. LLMs possess powerful capabilities in multimodal understanding, generation and reasoning. They can accurately perceive the environment, interpret contextual information and user intents, and extract key semantic features that are most relevant to the task. Furthermore, through prompt engineering and in-context learning, LLMs can swiftly and effectively adapt to wireless communication environments and resource constraints, enabling intelligent and adaptive resource allocation.

%Multimodal large language models (MLLMs) present unprecedented opportunities to address the aforementioned challenges \cite{MLLM}, given their advanced capabilities in multimodal reasoning, contextual comprehension, and decision-making abilities. Integrating MLLMs into semantic communication systems is expected to achieve context-aware information prioritization, task-oriented bandwidth allocation, and intelligent task processing, thereby significantly enhancing both communication efficiency and task performance.

\begin{figure*}[!t]
\centering
\includegraphics[width=0.9\textwidth]{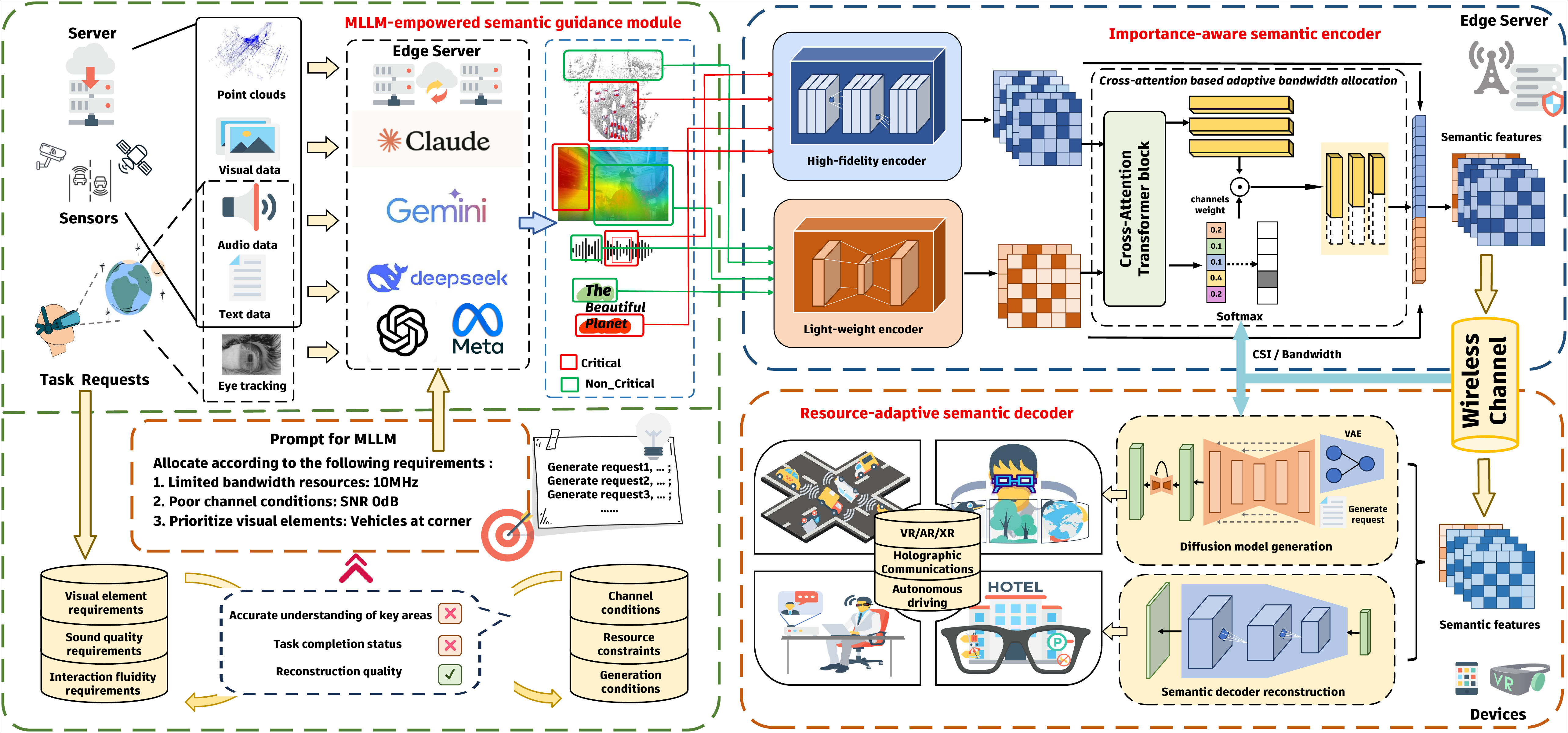}
\vspace{-2mm}
\caption{The proposed MLLM-SC framework. }
\label{Fig:MLLM-SC}
\end{figure*}

Recent studies have already started to employ LLMs for semantic communications. An LLM-based semantic communication framework specifically for image data \cite{Jiang-LLMSC-mag1} utilizes the segment anything model for semantic segmentation, and proposes attention-based semantic integration and adaptive semantic compression. Multimodal frameworks \cite{Jiang-LLMSC-mag2} employ MLLM-based alignment to unify all modalities into text format with personalized knowledge bases. Understanding-level semantic communication systems \cite{Guo-TMLCN-2025} transform visual data into human-intelligible content using image caption networks with LLM-based importance quantification and error correction. LaMoSC \cite{Zhao-TCCN-2024} integrates visual and textual features through attention mechanisms for image reconstruction.

However, deep integration of MLLMs and semantic communications remains to be explored. On the one hand, introducing MLLMs necessitates a redesign of the semantic encoder and decoder. For example, to effectively exploit semantic guidance from MLLM, the encoder should enable fine-grained control over compression fidelity, adaptively allocating more bandwidth to critical content while compressing less relevant regions more coarsely. The decoder should be capable of reconstructing or generating content from the received semantic features over varying channel conditions based on task prompts and local context. On the other hand, MLLMs can comprehend more than just the input data. By jointly providing task requirements, resource constraints, and channel conditions, they can generate more efficient semantic guidance.

This article presents a novel MLLM-integrated semantic communications framework, named MLLM-SC, within the device-edge collaborative architecture for immersive communications. The goal is to fully leverage pre-trained MLLMs to enhance wireless communication performance under time-varying channels and various resource constraints. First, we propose an MLLM-empowered semantic guidance module to comprehensively understand multimodal inputs, contexts, wireless channels and task requests, and export semantic guidance through attention heatmaps or masks to prioritize task-relevant information. Then, we design an importance-aware semantic encoder with dual-path architecture and cross-attention mechanism, which dynamically allocates bandwidth according to semantic guidance and channel states. Finally, we design a resource-adaptive semantic decoder based on variational autoencoders and conditional diffusion, which can consistently provide high-quality reconstruction and generation under varying bandwidths, channel conditions, and computing power. Case studies on MLLM-enhanced semantic communication and diffusion-driven image generation are conducted to showcase the effectiveness of the proposed framework.

%MLLM is deployed at edge to comprehensively understand multimodal input data, contexts, wireless channels and task requests, and export semantic guidance through attention heatmaps or masks to prioritize task-relevant information. Then, we propose a dual-path encoding architecture with cross-attention mechanisms that dynamically allocates transmission resources based on the semantic guidance. Variational autoencoders and conditional diffusion models are deployed at the device, which are robust to bandwidth constraints and channel conditions, and can reconstruct or generate content with high quality. Two case studies on MLLM-enhanced importance-aware semantic communication and diffusion-driven image generation are carried out to demonstrate the proposed framework.

\section{MLLM-Integrated Semantic Communications Framework}

% （介绍系统架构）
The proposed MLLM-SC framework is illustrated in Fig. \ref{Fig:MLLM-SC}. Task requests from the device side, such as eye-tracking signals and questions in AR/VR/XR, or the driving intent of vehicles, are transmitted to the edge server. The MLLM at the edge analyzes task requests and multimodal sensory inputs to understand user intent and the contextual scene. Based on this understanding, it analyzes high-dimensional data sources, such as 3D point clouds or high-resolution images or videos, to extract the semantics most relevant to the task. For example, in autonomous driving, the edge server can supplement vehicle perception by providing critical information from occluded or blind areas. The MLLM-empowered semantic guidance module integrates channel state information (CSI), resource constraints, and task requests to generate adaptive semantic instructions. By leveraging techniques such as prompt engineering, in-context learning, and soft prompting, the module enables resource-aware and task-specific conditioning of the pre-trained MLLM. The resulting guidance, typically in the form of attention heatmaps or binary masks, indicates the relative importance of different data segments. This guidance is incorporated into the semantic encoder to enable task-oriented, importance-aware, and fine-grained data transmission, improving overall communication efficiency.

%The MLLM then generates semantic guidance in the form of attention heatmaps or binary masks, which indicate the relative importance of different data segments. This guidance is incorporated into the semantic encoder to enable task-oriented, importance-aware and fine-grained data transmission, thus improving communication efficiency.

The semantic encoder dynamically allocates transmission resources based on task-oriented semantic guidance derived from MLLM, bandwidth constraints and channel conditions. Specifically, the data segments identified as critical by the MLLM are encoded with higher fidelity. More bandwidth is allocated to preserve the essential information, while background or less relevant segments are subject to coarse-grained encoding or aggressive compression. This selective treatment substantially reduces the transmission cost without compromising task-specific performance. On the device, the semantic decoder integrates the received information with locally available context to reconstruct original signals or generate new content. It supports bandwidth-aware reconstruction, adaptively adjusting the reconstruction fidelity based on channel conditions and dynamically synthesizing content in response to specific task requests or prompts, thereby enhancing flexibility under resource constraints.

%The semantic communication module deeply interacts with the MLLM by feeding back channel state information (CSI), resource constraints and task requests, enabling the MLLM to generate more accurate semantic guidance and further improve system performance. To fully exploit the capabilities of pre-trained MLLMs, we employ prompt engineering, in-context learning and soft prompting techniques to enable resource-adaptive and task-specific guidance through flexible and lightweight model conditioning.

% This feedback enables a closed-loop optimization framework, where the LLM is fine-tuned via reinforcement learning to better adapt its semantic prior generation strategies to varying channel conditions and task requirements. In addition, prompt engineering techniques are employed to refine the interaction between the device and the LLM, enabling more precise guidance in identifying and extracting task-relevant semantic content. Through this adaptive and feedback-driven process, the system achieves more intelligent and context-aware allocation of bandwidth and encoding resources, thereby improving the overall efficiency, adaptability, and robustness of semantic communication under real-world constraints.

\subsection{Typical Use Cases}

The proposed MLLM-SC framework is promising in various scenarios involving large-scale, high-dimensional data, where end devices are constrained by limited storage, computing power, and global contextual awareness. Typical applications include: 
%AR/VR/XR, high-definition and panoramic video streaming, LiDAR-based perception, and holographic communications. In such applications, the device requests task-relevant information from the edge server, which may store or relay high-dimensional data, or provide a more comprehensive understanding of the environment. Leveraging MLLM inference of user intent and context, the edge performs targeted semantic delivery that minimizes communication overhead and supports high-quality, low-latency experiences on resource-constrained devices.

 (1) \textit{Immersive AR/VR/XR Experiences in Smart City}. The goal is to provide immersive experiences by seamlessly integrating digital information with the physical environment in real time. As users navigate urban environments through lightweight AR/VR devices, the system monitors positional changes and viewing directions to predict required data segments. When user moves or issues visual queries about their surroundings, the device initiates targeted queries to the edge. These queries retrieve essential AR/VR data corresponding to current field of view, such as 3D building models and real-time traffic. The MLLM analyzes user intent to prioritize semantically relevant urban elements, such as navigation landmarks and safety-critical information, enabling smooth rendering of complex urban environments on resource-limited devices, and supporting intelligent visual question answering (VQA) capabilities.

  %Mobile device storage limitations necessitate intelligent edge-cloud collaboration for dynamic data loading.
    
 (2) \textit{Holographic Communications for Immersive Conferencing}. Next-generation holographic conferencing requires real-time transmission of high-fidelity 3D representations for natural interaction. The massive data requirements of holographic content pose significant transmission challenges over bandwidth-constrained networks. The MLLM analyzes participant gestures, facial expressions, voice patterns, and eye-tracking to understand communication intent and emotional context, while also processing visual queries about meeting content. Based on semantic understanding, the system dynamically allocates resources to prioritize perceptually important regions, such as facial features during emotional expressions or hand gestures during presentations. The importance-aware framework selectively compresses background elements while preserving high-fidelity details for critical content, enabling realistic holographic presence and intelligent content understanding while dramatically reducing bandwidth requirements.

  (3) \textit{Autonomous Driving with Vehicle-Edge Collaboration}. 
 In connected autonomous driving, the edge infrastructure delivers global environmental awareness to compensate for vehicle occlusions. Through the proposed MLLM-SC framework, the MLLM processes multimodal inputs from road-side cameras, LiDARs, and queries of vehicles to understand the complex traffic scenes and reason about vehicle data requirements. Then, the MLLM provides guidance to the semantic communication module to prioritize encoding and transmission of multimodal sensor data in critical occluded areas. Highly important safety information receives dedicated high-reliability channels, while non-occluded data undergoes bandwidth-efficient compression, ensuring life-critical information reaches autonomous vehicles with minimal latency through intelligent scene understanding and visual reasoning capabilities.

\subsection{Key Challenges}
%\textcolor{blue}{GuoLei, list three-four major challenges, 300-400 words}

%Several core challenges remain to be addressed for practical deployment:

(1) \textit{Efficient Semantic Alignment Across Heterogeneous Modalities.}
In multimodal communication scenarios such as human-machine interaction and autonomous systems, diverse task requests, including eye-tracking signals, voice and textual commands, can be jointly interpreted to support task-oriented transmission. We adopt semantic alignment techniques that project heterogeneous modalities into a shared representation space like Contrastive Language-Image Pre-training (CLIP) \cite{radford2021learning}, which align visual and textual data through contrastive learning. Effective mapping and alignment within a shared semantic space is fundamental to supporting downstream inference and task-oriented transmission. 

%Although state-of-the-art LLMs possess certain cross-modal understanding capabilities, they currently lack consistent modeling strategies for dynamic environments, especially regarding modality collaboration and spatial-semantic mapping. Moreover, robust alignment mechanisms are still needed to bridge the cross-modal relationships between resource-constrained device inputs and massive external data at the edge, such as dense point clouds and high-resolution video. 

(2) \textit{Semantic Importance-Driven Dynamic Communication Resource Allocation.}
In bandwidth-limited and latency-sensitive scenarios, resource allocation strategies must closely reflect task semantics to enhance overall transmission efficiency and response quality. Traditional scheduling schemes primarily rely on static network-level metrics, failing to capture task priority and semantic saliency adequately. The proposed architecture introduces a semantic-aware resource allocation mechanism guided by attention heatmaps, enabling differentiated treatment of critical versus non-critical regions. Nonetheless, challenges remain in refining resource partition granularity, adapting to time-varying channels, and establishing robust closed-loop feedback control.

(3) \textit{Semantic Content Reconstruction and Generation Under Resource Constraints.}
In generative communication tasks, devices are constrained by limited computation, memory, and energy, which restricts their ability to perform high-quality image reconstruction or semantic recovery locally. To address this, we propose integrating MLLMs and diffusion models with a variational autoencoder (VAE)-based compression and reconstruction network, enabling task-oriented content generation with minimal transmission overhead. In this framework, critical semantic features are extracted and transmitted to the device, where generative models reconstruct content based on both the received semantic and locally available contextual information. This collaborative architecture reduces the bandwidth requirement, and also enhances generation quality by using powerful pre-trained MLLMs and diffusion models. 

%(4) \textit{Reinforcement Learning Framework for Cross-Module Coordination.}
%The system’s various functional modules—including semantic intent parsing, resource allocation policy generation, semantic guidance creation, edge-side reconstruction, and feedback—are highly interdependent. Developing a unified reinforcement learning framework that models channel states, task feedback, and generation quality is crucial for dynamic, system-level optimization. However, significant challenges persist in designing appropriate reward functions, state representations, and multi-policy joint training strategies. In particular, ensuring training stability and generalization under high-dimensional state spaces and delayed feedback remains a key research focus. \cite{GL-TWC} \cite{GJY-JSAC}

\begin{figure*}[!t]
\centering
% Main system architecture
\includegraphics[width=0.85\textwidth]{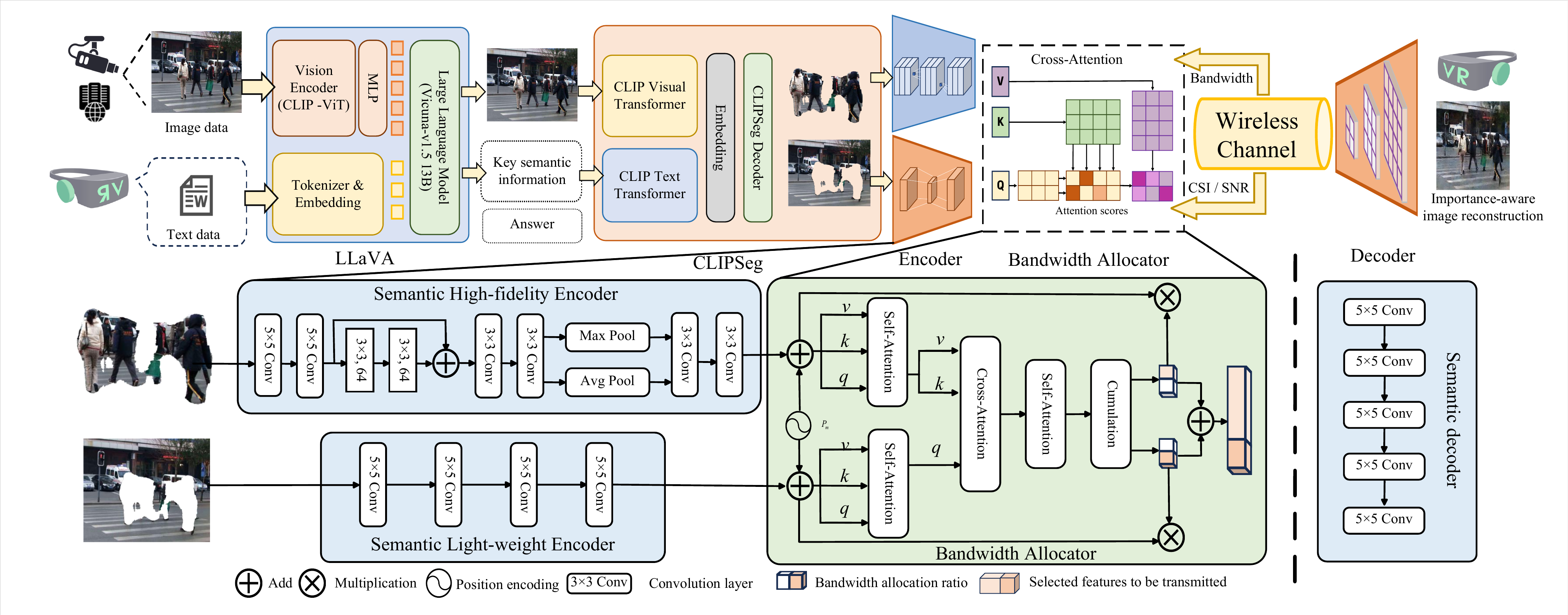}
\vspace{-2mm}
\caption{MLLM-enabled importance-aware semantic communication system architecture for VQA tasks.}
\label{Fig:MLLM-SC-VQA}
\end{figure*}

\subsection{Core Components}
%textcolor{blue}{Yusong, 600 words}

The MLLM-SC system comprises three core modules that collaboratively enable efficient multimodal intelligent communication, as illustrated in Fig. \ref{Fig:MLLM-SC}. The overall system architecture demonstrates the complete pipeline from multimodal inputs to adaptive semantic reconstruction or generation across diverse applications.

(1) \textit{MLLM-Empowered Semantic Guidance Module}
 
 This module leverages pre-trained MLLMs to process multimodal inputs and task requests, and generates semantic guidance for downstream processing. Task requests include eye-tracking signals from AR/VR applications, driving intent queries from vehicles, and natural language commands from interactive systems. Wireless channel conditions in the form of CSI or signal-to-noise ratio (SNR), along with available bandwidth and computing power are also given to the MLLM for generating channel- and resource- adaptive semantic guidance. Note that rather than fine-tuning LLMs, our goal is to harness the capabilities of pre-trained models by designing a semantic communication architecture tailored to them.
 
 State-of-the-art MLLMs are leveraged to perform cross-modal understanding and semantic reasoning. Representative MLLMs include GPT-4V/GPT-4o for robust visual understanding and Qwen2.5-VL for multilingual visual reasoning. For multimodal alignment, the system employs CLIP as a fundamental component to bridge vision and language modalities via contrastive learning within a shared embedding space. Inspired by CLIP, CLAP extends this framework to explore alignment between audio and language modalities, enabling contrastive pretraining across speech and text domains. BLIP-2 is incorporated with its Q-Former architecture for enhanced cross-modal feature extraction and bootstrapping mechanism for filtering noisy web data \cite{BLIP2}. Additionally, ALIGN provides large-scale contrastive learning capabilities for robust zero-shot performance across diverse domains. 
 %\cite{10480348}
 
 The semantic guidance process generates importance-aware attention maps that differentiate between key scenario areas and secondary scenario areas, as visualized in Fig. \ref{Fig:MLLM-SC}. CLIPSeg \cite{CLIPSeg} generates precise binary masks through text-conditioned semantic segmentation, producing spatially-oriented attention heatmaps corresponding to specific text prompts. The penultimate layer embeddings from CLIPSeg represent fine-grained importance maps that align with semantic queries. This differentiated semantic understanding enables the system to identify which regions or content elements require high-fidelity transmission versus those that can tolerate more aggressive compression. The module incorporates a closed-loop feedback optimization mechanism, continuously collecting feedback from task requests, channel conditions, and resource constraints to refine its semantic understanding strategies.

    %The MLLM at the edge or device analyzes the user's environment and their multimodal inputs (speech, gestures, context). This allows the system to grasp the user's intent and the relevant aspects of their surroundings. %体现出无线传输对大模型的反向指导
    %\textcolor{red}{multimodality, CLIP}
 (2) \textit{Importance-aware Semantic Encoder Module}
 
Building upon the semantic guidance from the MLLM module, the importance-aware semantic communication module implements differentiated processing and adaptive transmission strategies. As illustrated in the upper-right portion of Fig.  \ref{Fig:MLLM-SC}, this module features a dual-path architecture with both high-fidelity and lightweight semantic encoder networks. The high-fidelity network handles semantically important regions identified by the guidance module, employing deeper network architectures to preserve critical visual details with higher fidelity. Simultaneously, the lightweight network processes secondary scenario areas using lightweight architectures and more aggressive compression strategies to reduce transmission overhead. To balance semantic representation, noise resistance, and computational efficiency, both encoders can be equipped with a policy network that leverages differing capabilities of deep and shallow networks, enabling adaptive depth selection based on content complexity and resource constraints \cite{10772628}.
 
 The module employs cross-attention mechanisms to compute query (Q), key (K), and value (V) matrices, generating attention scores that reflect the relative importance of different semantic regions. These attention scores drive the bandwidth allocation process, where high-importance features receive substantially more transmission resources compared to secondary features with lower priority scores. The adaptive transmission component implements dynamic resource scheduling based on channel conditions. This semantic importance-aware resource allocation ensures that even under deteriorating channel conditions, the most important content maintains transmission quality while secondary regions sacrifice fidelity to preserve overall system performance.

 (3) \textit{Resource-adaptive Semantic Decoder Module}
 
 The resource-adaptive semantic decoder module aims to dynamically adapt its reconstruction and generation strategies based on device computing capabilities, available bandwidth, and current channel conditions. 
 %As shown in the right portion of Fig. \ref{Fig:MLLM-SC}, this module supports diverse application scenarios including VR/AR/XR experiences, holographic communications, and autonomous driving systems. 
 The module leverages advanced conditional diffusion models to perform a multi-step denoising process guided by device-side prompts or contextual task requests, enabling high-quality content generation even under low bandwidth conditions. To further reduce transmission bandwidth, following the generative step, the module employs VAE techniques to decode and reconstruct compressed semantic features from the received signals, ensuring semantic consistency and structural integrity in the multimodal content. The decision reasoning component implements comprehensive optimization processes that consider task requirements, resource constraints, and quality targets. Resource adaptation is achieved through comprehensive decision reasoning mechanisms that consider bandwidth-aware processing, terminal computing adaptation, and channel condition responsiveness. The system dynamically scales processing complexity according to available resources while maintaining acceptable quality levels for critical content elements. The diffusion-based generation process enables the system to synthesize high-quality content even when only partial semantic information is transmitted, significantly improving bandwidth efficiency for immersive applications. 
 
 %Through these three modules, the proposed architecture exploits the reasoning and generative capabilities of MLLM, achieves end-to-end optimization for task-oriented and importance-aware semantic communications, thus enabling efficient and robust communication for immersive experiences.

\subsection{Performance Metrics}
%\textcolor{blue}{Loss function design? Overall training strategies? Latency?}
%\textbf{Weighted Mean Squared Error for Importance-Aware Reconstruction}
%For Case Study I focusing on visual question answering, the system implements a Weighted MSE loss function that emphasizes semantically critical regions identified by the MLLM guidance module. The loss function assigns exponentially higher weights to masked regions containing question-relevant content, enabling the network to prioritize accurate reconstruction of areas essential for answer generation. This differentiated weighting strategy ensures that bandwidth-limited transmission preserves the most important visual information while allowing aggressive compression of background regions.

%\textbf{VAE-Based Distribution Matching Loss for Generative Communications}
%Case Study II employs a comprehensive VAE-based loss function combining reconstruction fidelity and distribution alignment objectives. The training objective jointly optimizes VAE reconstruction loss and KL-divergence-based guidance loss to ensure that compressed semantic features maintain the Gaussian distribution required by diffusion models. This approach enables robust content generation under bandwidth constraints while preserving semantic coherence between transmitted features and reconstructed outputs.

To comprehensively evaluate the performance of the proposed semantic communication system, we adopt a range of metrics covering pixel-level fidelity, semantic-level perception, and potential task-level relevance.

Pixel-level metrics such as peak signal-to-noise ratio (PSNR) and structural similarity index measure (SSIM) assess the visual fidelity of reconstructed images. PSNR focuses on pixel-wise reconstruction errors, while SSIM evaluates structural similarity based on luminance, contrast, and texture. These metrics primarily reflect low-level details and are suitable for measuring distortion.

Semantic-level metrics focus on whether the essential content and meaning of the transmitted data are preserved. CLIP-score quantifies semantic similarity using a vision-language model, comparing the reconstructed image to the reference based on their embeddings. Frechet inception distance (FID) evaluates the distributional distance between generated and real images in a deep feature space, reflecting the realism of the output. Learned perceptual image patch similarity (LPIPS) measures perceptual similarity from a human-centric perspective using deep neural features. In this context, higher CLIP-score and lower FID and LPIPS values indicate better semantic transmission performance.

Task-level metrics evaluate whether the transmitted content preserves sufficient semantic information to support downstream tasks such as classification, detection, or segmentation. These metrics reflect the practical utility of semantic communication in real-world applications. Common examples include accuracy for classification, mean average precision (mAP) for object detection, and intersection over union (IoU) for segmentation. They provide a direct measure the performance of the transmitted data to support task-specific performance.

\section{Case Study I: MLLM-Enhanced Importance-Aware Semantic Communications}

% In 6G networks supporting autonomous driving, edge-vehicle collaboration for real-time scene understanding faces significant bandwidth challenges for high-resolution image transmission. This case study demonstrates intelligent visual question answering (VQA) services for traffic scenarios over bandwidth-constrained wireless channels.

% Autonomous vehicles capture environmental images and pose natural language questions about safety-critical regions. For instance, vehicles might inquire: ``Are there pedestrians crossing in the intersection ahead?'' or ``What is the status of the traffic light in this image?'' Such queries require powerful visual understanding capabilities and accurate parsing of safety-critical question semantics for precise visual-linguistic associations.

% Traditional complete image transmission is inefficient when vehicle queries focus on specific safety-critical regions, consuming valuable bandwidth on irrelevant visual information. Autonomous driving VQA tasks demand systems capable of intelligently identifying and prioritizing image regions semantically relevant to safety-critical questions.

%场景修改为在智慧城市的沉浸式AR/VR
This case study targets AR/VR applications and demonstrates intelligent VQA services, where users can pose natural language queries about their surroundings to edge servers over bandwidth-constrained wireless channels. AR/VR users capture urban environmental scenes and issue visual queries about urban elements they encounter. For instance, users might inquire: ``What is that building across the street?'' or ``Are there any restaurants nearby?'' Such queries require powerful visual understanding capabilities and accurate parsing of contextual question semantics for precise visual-linguistic associations.
Traditional image transmission is inefficient when user queries focus on specific regions of interest. 
%Instead, the communication system needs to intelligently identify and prioritize regions semantically relevant to user questions, thereby enabling seamless integration of digital information with the physical environment.

\begin{figure}[!t]
\centering
\includegraphics[width=0.95\columnwidth]{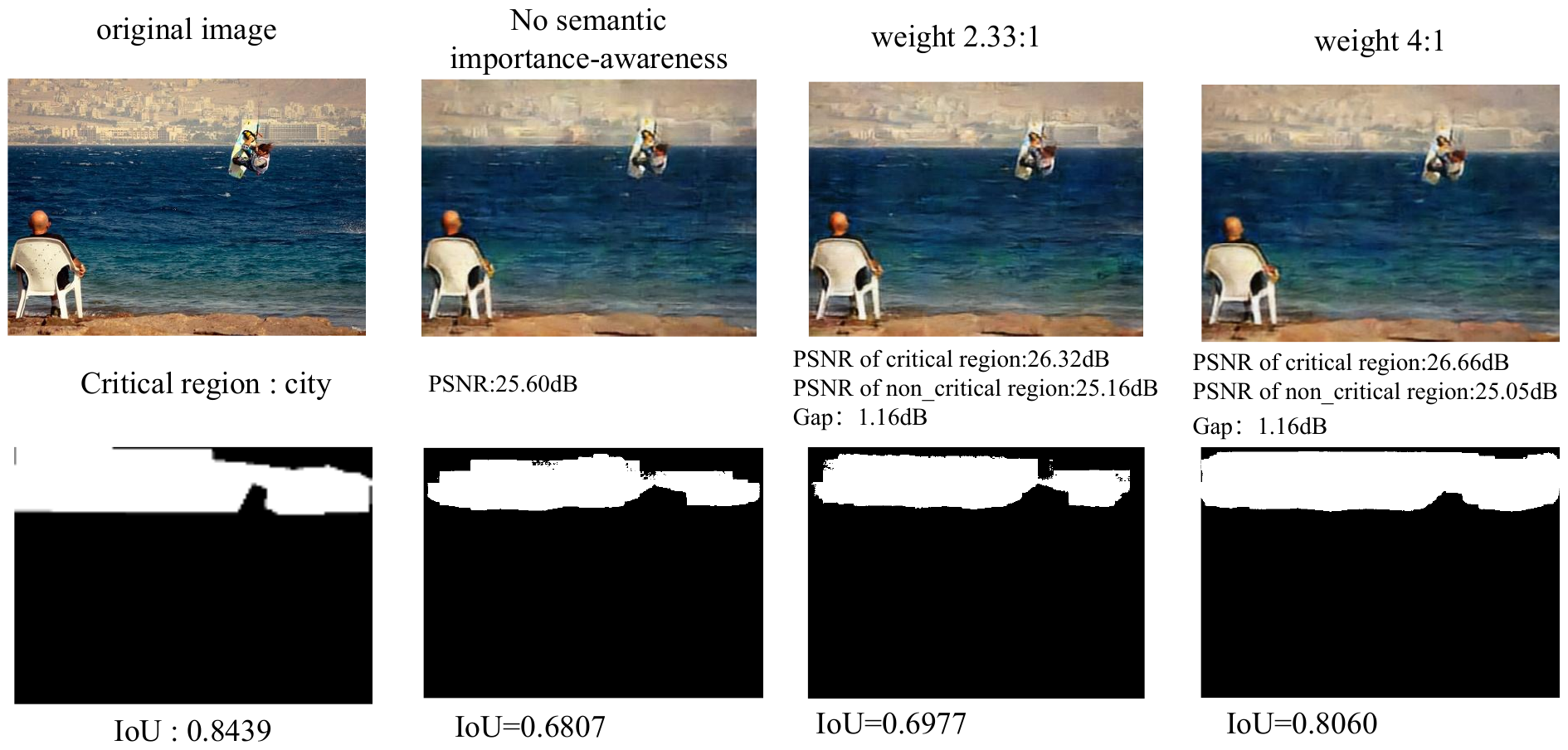}
\vspace{-2mm}
\caption{Visual comparisons with or without MLLM under different importance weights.}
\label{fig:weighting_comparison}
\end{figure}

The workflow of our MLLM-enhanced importance-aware semantic communication system is illustrated in Fig.~\ref{Fig:MLLM-SC-VQA}. 
%Our VQA implementation leverages the three-module architecture with specific adaptations for question-answering scenarios. The system processes dual inputs: high-resolution environmental imagery and corresponding natural language questions that require visual reasoning and semantic analysis.
 The semantic guidance module employs LLaVA \cite{LLaVA-1.5} as the core multimodal reasoning engine, combining CLIP ViT-L visual encoding with Vicuna-v1.5 13B language modeling for sophisticated question-image understanding. The CLIPSeg \cite{CLIPSeg} component performs question-guided region segmentation, generating binary masks that identify image areas relevant to posed questions through semantic matching in CLIP embedding space.

The importance-aware communication module implements dual-path encoding architecture. 
%High-fidelity network processing handles question-relevant masked regions with high-fidelity treatment, while lightweight encoding compresses background regions with lightweight architectures. This differentiated approach enables substantial bandwidth savings without compromising answer accuracy.
Cross-attention based bandwidth allocation module adaptively allocates available bandwidth to different regions according to question complexity and visual reasoning requirements. Simple factual questions about object attributes receive focused allocation to relevant regions, while complex reasoning questions requiring broader scene understanding maintain higher background quality.

%prioritizing regions that contribute to answer accuracy. The system analyzes question complexity and visual reasoning requirements to determine optimal resource distribution. Simple factual questions about object attributes receive focused allocation to relevant regions, while complex reasoning questions requiring broader scene understanding maintain higher background quality.

We use VGPhraseCut dataset containing 345,486 phrase-region pairs across 77,262 images for experiments. We selected samples with masks occupying 10-40 ratio of total image area, aligning with typical user attention distribution in practical VQA applications. The training explicitly distinguishes between mask regions (semantically critical areas) and non-mask regions through weighted mean square error (MSE) loss. For evaluation, we employ PSNR for reconstruction quality and IoU for semantic preservation. The IoU evaluation follows a systematic process where original and transmitted images undergo CLIPSeg segmentation, with resulting masks compared against ground truth to compute the IoU.

The visualization results are provided in Fig. \ref{fig:weighting_comparison}. To demonstrate the effectiveness of our framework, we use a representative VQA example, where the user asks ``What is across the sea?" with the ground truth answer being ``city". In this scenario, the MLLM identifies the city region across the sea as semantically critical for answering the question, while the sea and other background areas are considered less important. The importance weight represents the bandwidth allocation ratio between critical regions identified by MLLM masks and background regions during transmission. The baseline semantic communication method without MLLM guidance and importance-awareness achieves balanced reconstruction but suffers substantial semantic degradation, demonstrating the limitations of uniform bandwidth allocation. As the importance weight increases to 2.33:1 with MLLM-guided semantic awareness, semantic preservation improves notably. With weight 4:1, the MLLM-enhanced system demonstrates significant semantic preservation improvements with transmission IoU of 0.8060 and IoU degeneration of only 0.0279. Higher weighting ratios enabled by MLLM guidance maintain better semantic integrity of critical regions, particularly preserving essential objects for VQA task performance, highlighting the effectiveness of importance-aware resource allocation.

%The differentiated weighting enables the network to prioritize accurate reconstruction of semantically critical regions during training, directly simulating importance-driven resource allocation during transmission. High-importance regions receive more bandwidth and encoding resources, while low-importance regions undergo aggressive compression, ensuring learned representations align with the importance-aware semantic communication framework.

%Fig. \ref{fig:weighting_comparison} shows the effectiveness of importance-aware weighting strategy through comparative analysis of different weight allocation schemes using PSNR for reconstruction quality and IoU for semantic preservation. The IoU evaluation follows a systematic process where original and transmitted images undergo CLIPSeg \cite{CLIPSeg} segmentation, with resulting masks compared against ground truth to compute transmission IoU scores.

\begin{figure}[!t]
\centering
\includegraphics[width=\linewidth]{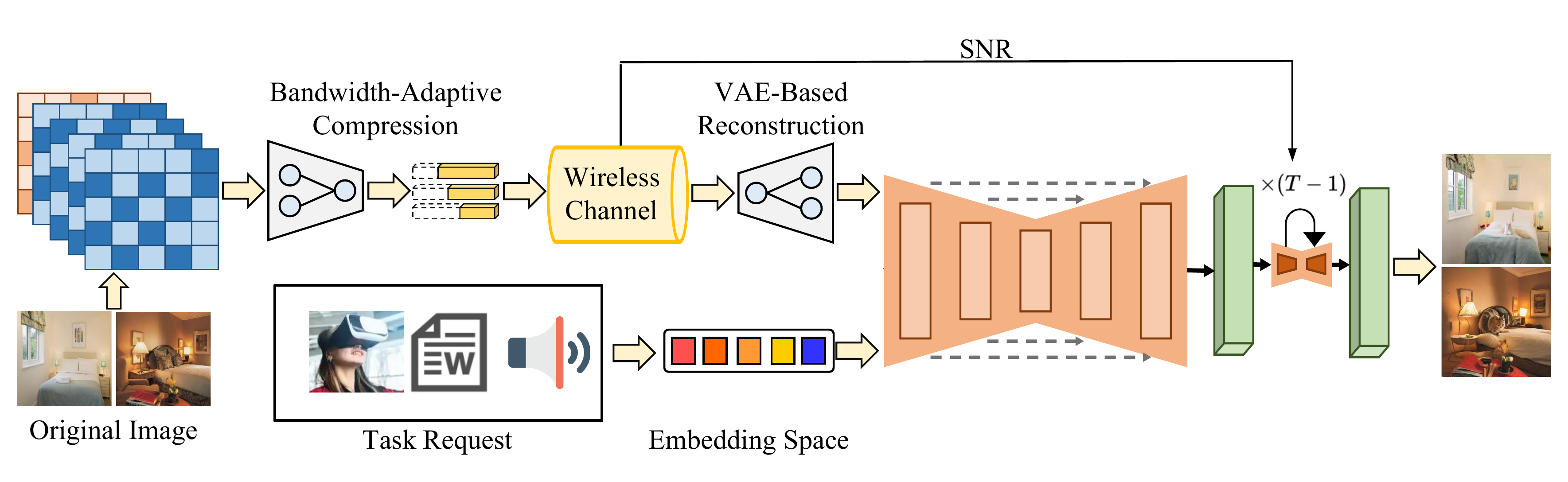}
%\vspace{-5mm}
\caption{Architecture of the diffusion-driven image generation in MLLM-enabled semantic communication system.}
\label{Fig:MLLM-SC-diffusion}
\end{figure}

\section{Case Study II: Diffusion-Driven Semantic Communications for Image Generation}
%\textcolor{blue}{GuoLei, about 1000 words, 1 figure of system architecture, 1 figure of simulation results}

%（先介绍生成的目的，在介绍困难之处，最后解决方法 prompt + no prompt，实验结果为线图和可视化结果）

%In this section, we propose a case study that applies the diffusion-driven semantic communication integrated with bandwidth-limited compression. As illustrated in Fig. \ref{Fig:MLLM-SC-diffusion}, this framework demonstrates its great potential in enhancing generative semantic communication performance under stringent wireless bandwidth constraints.

In this section, we present a case study on a diffusion-driven semantic communication framework under bandwidth-limited conditions, which builds upon our previous work \cite{GL-TWC}. Diffusion is employed as a generative decoder guided by MLLMs, and further supports prompt-based inputs to enable controllable generation. This integration combines the semantic understanding capability of MLLMs with the generative capability of diffusion models to achieve high-quality reconstruction or genration. As shown in Fig. \ref{Fig:MLLM-SC-diffusion}, the framework operates on the semantic recovery of the system, where compressed features transmitted over bandwidth-constrained channels are decoded on the device.

%In this section, we present a case study on a diffusion-driven semantic communication framework under bandwidth-limited conditions, extending our previous work \cite{GL-TWC}. In this framework, diffusion models act as generative decoders guided by multimodal large language models (MLLMs), which provide high-level semantic priors to enhance reconstruction quality. This integration combines the semantic understanding capability of MLLMs with the fine-grained generation capacity of diffusion models to achieve high-quality signal reconstruction under strict bandwidth constraints. As illustrated in Fig. \ref{Fig:MLLM-SC-diffusion}, the framework operates on the semantic recovery of the system, where compressed features transmitted over bandwidth-constrained channels are decoded and transformed into task-relevant outputs at the device.

%This design leverages the powerful generative capacity of diffusion models to achieve high-quality semantic reconstruction in low-resource wireless environments.

To enable generative semantic communication in over limited bandwidth, we design a transmitter that encodes the semantic features of a source image and transmits them over a noisy wireless channel. The key innovation lies in modeling the wireless transmission as the forward process of a diffusion model, thereby allowing the receiver to leverage the reverse diffusion process (e.g., stable diffusion) for noise elimination and content reconstruction. To mitigate the bandwidth overhead of transmitting full-resolution latent features, a compression mechanism is introduced to preserve essential semantics while significantly reducing transmission overhead.

At the receiver, considering that the compression and reconstruction modules can disrupt the original Gaussianity of latent features, a VAE-based reconstruction module is employed. This module upsamples the compressed semantic features and reconstructs them, ensuring that the recovered features conform to the Gaussian distribution required by diffusion models via reparameterization. To enhance robustness under varying channel conditions, the VAE decoder is further conditioned on the SNR, allowing adaptive estimation of feature variance.

To align the distribution of the reconstructed features with that of the original uncompressed data, a distribution matching strategy is introduced by minimizing the divergence between the VAE-derived posterior and the distribution of the original generator. The overall training objective jointly optimizes two components: the VAE reconstruction loss $L_{\text{VAE}}$, which ensures accurate semantic recovery, and a KL-divergence-based guidance loss $L_{\text{g}}$, which enforces distribution alignment. This joint optimization enables the model to maintain semantic fidelity under bandwidth constraints.

For simulations, the proposed framework is evaluated under additive white gaussian noise (AWGN) channel. The experiments focus on image generation tasks using the LSUN-Bedrooms datasets. Test SNR levels range from 5 dB to 12 dB. During training, the backbone diffusion model (e.g., stable diffusion) is kept frozen, while only the compression module and the VAE-based reconstruction module are updated under bandwidth constraints. The benchmark method is channel denoising diffusion model (CDDM) \cite{CDDM}, which combines diffusion generation with DJSCC and is evaluated under the same compression ratios and SNR conditions as the proposed method for a fair comparison.

\begin{table}
\begin{center}
\caption{Performance Comparison on the LSUN-Bedroom Dataset.}
\vspace{-2mm}
\begin{tabular}{c|ccc|cccc}
\hline
\!\!\multirow{2}{*}{Metric}\!\!\!\! \!\!& \multicolumn{3}{c|}{Methods} & \multicolumn{4}{c}{SNR} \\ 
\cline{2-8}
 & \!\!VAE \!\!& \!\!$L_{\text{MSE}}$\!\! &\!\! $L_{\text{VAE}} \!\!+ \!\! L_{\text{g}}$\!\!\!\! &\!\!\!\!\!\! 5dB\!\! &\!\! 7dB \!\!& \!\!9dB\!\! & \!\!12dB\!\! \\ 
\hline
\!\!\multirow{3}{*}{PSNR $\uparrow$}\!\! &  & \!\!\!\!\!\! \checkmark (CDDM)\!\!\!\!\!\! & & 14.56 & 14.82 & 14.96 & 15.02 \\
 & \checkmark & \checkmark & & 18.32 & 18.72 & 18.94 & 19.09 \\
 & \checkmark & \checkmark & \checkmark & 19.45 & 20.15 & 20.55 & \textbf{20.76} \\
\hline
\!\!\multirow{3}{*}{SSIM $\uparrow$}\!\! & & \!\!\!\!\!\!\checkmark (CDDM)\!\!\!\! \!\!& & 0.35 & 0.36 & 0.37 & 0.37 \\
 & \checkmark & \checkmark & & 0.47 & 0.48 & 0.49 & 0.49 \\
 & \checkmark & \checkmark & \checkmark & 0.52 & 0.54 & 0.56 & \textbf{0.57} \\
\hline
\!\!\multirow{3}{*}{CLIP $\uparrow$}\!\! & & \!\!\!\!\!\!\checkmark (CDDM) \!\!\!\!\!\!& & 0.49 & 0.50 & 0.53 & 0.54 \\
 & \checkmark & \checkmark & & 0.52 & 0.53 & 0.56 & 0.58 \\
 & \checkmark & \checkmark & \checkmark & 0.61 & 0.65 & 0.70 & \textbf{0.74} \\
\hline
\!\!\multirow{3}{*}{LPIPS $\downarrow$} \!\!& &\!\!\!\!\!\! \checkmark (CDDM) \!\!\!\!\!\!& & 0.69 & 0.68 & 0.68 & 0.68 \\
 & \checkmark & \checkmark & & 0.69 & 0.68 & 0.68 & 0.67 \\
 & \checkmark & \checkmark & \checkmark & 0.47 & 0.45 & 0.43 & \textbf{0.42} \\
\hline
\end{tabular}
\label{tab:bedroom}
\end{center}
\end{table}

Table \ref{tab:bedroom} presents the evaluation results of the proposed framework in comparison with the baseline CDDM, under a compression ratio of 1.3\%. The metrics include PSNR and SSIM in pixel-level fidelity, as well as LPIPS and CLIP to evaluate semantic consistency. In this table, higher values of PSNR, SSIM, and CLIP (↑) indicate better performance, while lower LPIPS (↓) reflects improved quality.

Across all SNR levels, the proposed method consistently outperforms CDDM in both pixel-level and semantic-level metrics. For example, at 12 dB, it achieves a PSNR of 20.76 dB, significantly higher than the baseline (15.02 dB). SSIM is also improved to 0.57 compared to lower scores in baselines. At the semantic level, our method achieves a higher CLIP score of 0.74 and a lower LPIPS score of 0.42, indicating better semantic consistency and perceptual quality. Overall, the proposed framework effectively enhances reconstruction fidelity and semantic preservation, especially under limited bandwidth conditions.

%Across all settings, the proposed method consistently outperforms CDDM. For example, in the MIMO at SNR = 11 dB, our method achieves a PSNR of 21.6 dB, compared to approximately 20 dB for CDDM. A similar trend is observed in SSIM, where our approach yields higher structural similarity across channel types. In terms of semantic-level metrics, the proposed framework achieves a lower LPIPS score (e.g., 0.37 vs. 0.45) and a higher CLIP score (e.g., 0.75 vs. 0.73) than CDDM, indicating better alignment with the intended semantic content. These results highlight the effectiveness of the proposed method in preserving both visual fidelity and semantic meaning under diverse and bandwidth-constrained wireless communication scenarios.

\begin{figure}
\centering
\includegraphics[width=0.85\linewidth]{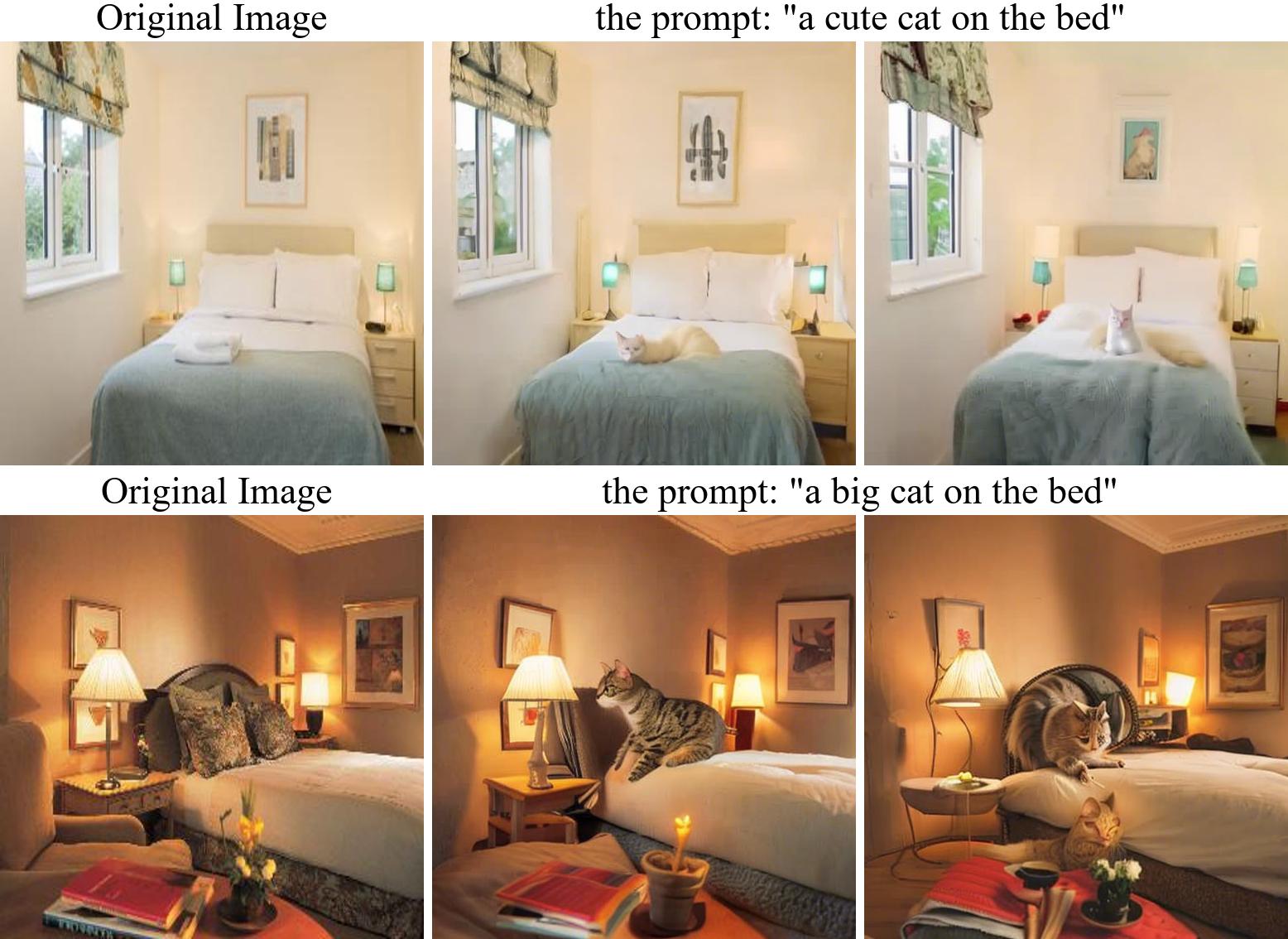}
\caption{Visual results for generation with textual prompts.}
\label{fig:system_with_prompt}
\end{figure}

% 介绍结果+举例子说明场景
Fig. \ref{fig:system_with_prompt} presents the visual results under the guidance of the prompts. Given an original image of a bedroom, the prompt ``a cute/big cat on the bed" is used to direct semantic generation at the receiver. The reconstructed images incorporate the target object while preserving the overall structure and style accordingly using the diffusion-based generative model. As shown in the two visualizations, the generative process effectively integrates textual guidance and reconstructs semantically aligned images, with a cat appearing on the bed. This illustrates the system capability to adaptively generate content based on prompts, which is particularly beneficial in resource-constrained scenarios such as AR/VR or situational content completion, where users can further refine or modify the received data according to specific task requirements.
 
%This case study demonstrates the effectiveness of combining diffusion-based generation with VAE-guided semantic compression for bandwidth-constrained communication. Extensive evaluations show that the proposed framework consistently outperforms baselines across diverse channel conditions in both pixel- and semantic-level metrics. Furthermore, prompt-driven visualizations highlight its ability to adaptively generate task-relevant content based on receiver-side instructions, supporting efficient and flexible communication in applications such as AR/VR and personalized content completion.

\section{Conclusion and Outlook}
%\textcolor{blue}{In this article...}

In this article, we have presented a novel MLLM-integrated semantic communications framework (MLLM-SC) for 6G immersive communication scenarios. The proposed framework leverages the powerful reasoning and generative capabilities of pre-trained MLLMs to enable intelligent, context-aware wireless communication through three novel components: MLLM-empowered semantic guidance module, importance-aware semantic encoder, and resource-adaptive semantic decoder.
The MLLM-SC framework demonstrates significant advantages in real-time environment and context understanding, and highly-efficient wireless transmission through importance-aware dynamic resource allocation. Through case studies on VQA for AR/VR applications and diffusion-driven image generation, we have shown substantial improvements in reconstruction quality and semantic preservation compared to conventional approaches. The experimental results have demonstrated that the integration of MLLMs with semantic communications represents a paradigm shift toward intelligent communication systems, enabling efficient and robust immersive experiences in resource-constrained 6G environments.

Future directions include enhancing the feedback-driven optimization framework by reinforcement learning methods that allow MLLMs to adaptively adjust generation strategies to varying channel conditions and task requirements. Real-time inference for MLLMs remains a major challenge, motivating research into model compression and efficient architectures to achieve low-latency deployment without sacrificing accuracy. Furthermore, multi-agent collaborative semantic communication promises improved resource utilization and robustness, with future efforts focusing on scalable coordination and adaptive resource management among distributed devices. These directions collectively aim to advance semantic communication toward practical, efficient, and intelligent real-world applications.

\end{document}